\pdfoutput=1
\documentclass[11pt]{article}

\usepackage[]{style/ACL2023}

\usepackage{times}
\usepackage{latexsym}

\usepackage[T1]{fontenc}
\usepackage[utf8]{inputenc}

\usepackage{microtype}
\usepackage{booktabs}
\usepackage{multirow}
\usepackage{amsmath}
\usepackage{amssymb}
\usepackage{graphicx}
\usepackage{balance}
\newcommand*{\scale}[2][4]{\scalebox{#1}{$#2$}}

\usepackage{pifont}% http://ctan.org/pkg/pifont
\newcommand{\cmark}{\ding{51}}%
\newcommand{\xmark}{\ding{55}}%

\usepackage[breaklinks]{hyperref}

\usepackage{inconsolata}

\newcommand{\model}{\textsc{LLM-URL} }

\title{Large Language Models are Built-in Autoregressive Search Engines}

\author{Noah Ziems, Wenhao Yu, Zhihan Zhang, Meng Jiang \\
  University of Notre Dame \\
  \texttt{\{nziems2, wyu1, zzhang23, mjiang2\}@nd.edu}
}

\begin{document}
\maketitle
\begin{abstract}
% Open Domain Question Answering (ODQA) is traditionally a two step process consisting of a \textit{retriever} and a \textit{reader}.
% Traditional approaches to ODQA retrieval 
% Dense retrievers such as DPR have shallow interactions and must be explicitly trained on gold-labeled question-document pairs.
% Recently, new approaches to retrieval have shown impressive performance by training a language model to directly generate document identifiers when given a question.
% In this work, we show existing Large Language Models (LLMs) such as GPT3 are already capable of generating document identifiers (URLs) without any explicit training.
% We show this framework, referred to as \model, consistently improves the recall@k of existing baseline retrieval methods by a significant margin on three different ODQA datasets: Natural Questions, TriviaQA, and WebQuestions.
% Further, we show \model improves downstream EM scores compared to existing methods.

Document retrieval is a key stage of standard Web search engines. 
Existing dual-encoder dense retrievers obtain representations for questions and documents independently, allowing for only shallow interactions between them. 
To overcome this limitation, recent autoregressive search engines replace the dual-encoder architecture by directly generating identifiers for relevant documents in the candidate pool.
However, the training cost of such autoregressive search engines rises sharply as the number of candidate documents increases.
In this paper, we find that large language models (LLMs) can follow human instructions to directly generate URLs for document retrieval.

Surprisingly, when providing a few {Query-URL} pairs as in-context demonstrations, LLMs can generate Web URLs where nearly 90\% of the corresponding documents contain correct answers to open-domain questions.
In this way, LLMs can be thought of as built-in search engines, since they have not been explicitly trained to map questions to document identifiers.
Experiments demonstrate that our method can consistently achieve better retrieval performance than existing retrieval approaches by a significant margin on three open-domain question answering benchmarks, under both zero and few-shot settings.
The code for this work can be found at \url{https://github.com/Ziems/llm-url}.
\end{abstract}

\vspace{0.1in}
\section{Introduction}
\label{sec:introduction}

\begin{figure}[t]
    \centering
    \includegraphics[width=0.5\textwidth]{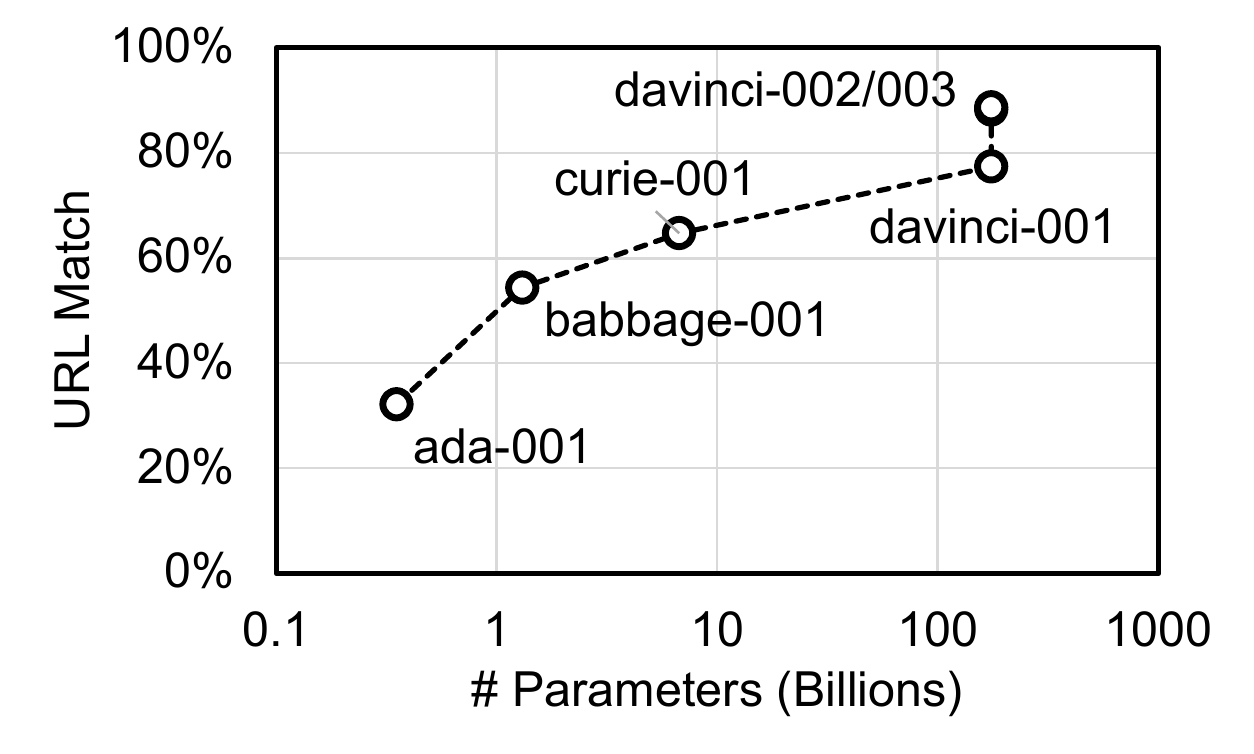}
    \vspace{-0.3in}
    \caption{Successful URL reconstructions by different size of GPT-3 as URL generators.
    The models are prompted with the first 100 words of a Wikipedia page then tasked with generating the URL of the page they came from. Tested on 10k Wikipedia pages sampled from the top 100k most frequent Wikipedia entities.}
    \label{fig:url-reconstructions}
    \vspace{-0.1in}
\end{figure}

Along with the success of deep learning, dual-encoder based retrievers have become the dominant method for Web searching~\cite{zhu2021retrieving,zhao2022dense}.
For example, DPR~\cite{karpukhin2020dense} employs two independent encoders to encode the question and the document respectively, then estimates their relevance by computing a single similarity score between two representations. 

However, these methods suffer from two major drawbacks.
First, the representations of questions and documents are typically obtained independently in modern dual-encoder dense retrieval models~\citep{karpukhin2020dense}, allowing for only shallow interactions between them~\citep{khattab2021relevance}.
Second, the question or document representation is embedded into a single dense vector, potentially missing fine-grained information when computing the similarity between the two vector representations~\citep{khattab2020colbert}.

% cover the benefits of NDI
Instead of computing similarity between question and document embeddings, autoregressive search engines aim to directly generate document identifiers then map them to complete documents in the predetermined candidate pool. This approach has attracted increasing interest in information retrieval (IR) and related fields~\cite{tay2022transformer,bevilacqua2022autoregressive,wang2022a}.
Compared to dual-encoder dense retrieval methods, autoregressive search engines enjoy a number of advantages.
First, autoregressive generation models produce document identifiers by performing deep token-level cross-attention, resulting in a better estimation than shallow interactions in dense retrievers.
Second, 
% The benefits of NDIs are numerous.
% Replacing shallow interactions with rich ones through many transformer layers leads to significantly better retrieval performance.
autoregressive search engines have been shown to have strong generalization abilities, outperforming BM25 in a zero-shot setting \cite{tay2022transformer}.
% discuss DSI with one sentence on DPR
% Where traditional dual encoder approaches such as Dense Passage Retriever (DPR) struggle from shallow interactions between query and document, new approaches have emerged to remedy this.
% One solution is to train a single Transformer \cite{vaswani2017attention} to map document passages and string queries to newly-created document ids, effectively embedding the contents of the entire corpus into the parameters of the Transformer.
% This approach, first introduced by \citet{tay2022transformer} as Differentiable Search Index (DSI), allows for rich interactions between document and query because the mapping is done by a large Transformer with many layers instead of a single dot product found with traditional approaches.
% In this work, we refer to this class of Transformers that map directly from passage/query to document id as \textit{neural document indexers} (NDIs).
% drawbacks of NDI
While it is theoretically possible to scale an autoregressive search engine to the size of a large language model (LLM), such as GPT-3 with 175B parameters, in practice it is not feasible due to the computational overhead of training such a large autoregressive search engine from scratch~\cite{tay2022transformer}.
%for some tasks~\cite{tay2022transformer}.
% Training DSI with only 1B parameters on all 320k questions in the Natural Questions dataset took over a day on a cluster of 256 TPUv4s.
% allude to NDIs not being capable of few shot
% In addition, adding new documents to the corpus or updating existing ones requires further computation, making fine tuning an NDI for downstream tasks difficult.
To reduce the high training cost of autoregressive search engine, a smaller model size is preferred.
However, the results of our pilot study in Figure~\ref{fig:url-reconstructions} show smaller language models are significantly worse at mapping passages to document identifiers than larger ones.
% and do not take advantage of NDIs scaling laws.
% why few-shot search engine is valuable
Moreover, different retrieval tasks can have unique retrieval requirements.
One task may require a model to retrieve factual evidence to support or refute a claim (\textit{i.e.}, fact checking)~\cite{CREAK} while another may require a model to retrieve specific trivia information about an entity (\textit{i.e.}, entity linking)~\cite{petroni2021kilt, EDMem}.
It would be better if the retriever was capable of generalizing to new retrieval tasks with only a few examples.

% Using pretrained LLMs avoid these problems
% Recently, a number of pre-trained large language models (LLMs) have emerged that achieve impressive performance on a wide range of NLP tasks~\cite{brown2020language,chowdhery2022palm}.
% Recent work has found that large language models have strong factual memory capabilities, and can directly generate supporting evidence for open-domain questions, thereby replacing retrievers~\cite{yu2022generate}.

% For tasks of factual text generation, a number of LLMs have shown an ability to generate citations with URLs for their text-generations \cite{bohnet2022attributed}.
% Though a small portion of URLs are hallucinated, more than 70\% of URLs are valid and contain the information sought by the question.
% In particular, our experiments show newer versions of these LLMs such as GPT-3's text-davinci-002/003 have shown strong abilities to generate URLs directly.

In this work, we explore the use of in-context demonstrations to prompt LLMs to directly generate web URLs for document retrieval, namely \textsc{LLM-URL}.
Surprisingly, we find that by providing a few (query, URL) pairs as contextual demonstrations, large language models (e.g. GPT-3) generate Web URLs where nearly 90\% of the corresponding documents contain answers to open-domain questions.
In this way, LLMs can be thought of as built-in search engines, as they have not been explicitly trained to map questions or documents to identifiers.
Instead of using newly-created document identifiers, \textsc{LLM-URL} leverages existing and widely used document identifiers directly, \textit{i.e.}, URLs.
We compare our approach to existing document retrieval methods on three different open-domain question answering (QA) datasets: WebQ~\citep{berant2013semantic}, NQ~\citep{kwiatkowski2019natural}, and TriviaQA~\citep{joshi2017triviaqa}.
% Further, to reduce the size and number of supporting passages, we investigate passage filtering using a simple ranking mechanism.
Further, to avoid exceeding the limit on the number of input tokens of LLMs, we employ an unsupervised passage filtering module to remove irrelevant portions of supporting documents.
To summarize, our main contributions are as follows:
\begin{enumerate}
  \item We reveal that LLMs are built-in autoregressive search engines capable of document retrieval by directly generating Web page URLs under both zero and few-shot settings.
  \item We show retrieving documents by generating URLs with LLMs significantly outperforms existing methods for document retrieval, as measured by Recall@K.
  Further, we show that breaking the retrieved documents into passages then using a ranker to filter the passages significantly reduces the number of supporting passages while maintaining high recall.
  \item We show the retrieved documents improve downstream QA performance as measured by EM when compared to baseline methods.
  % \item We show that breaking the generated documents into passages then using a ranker to filter the passages significantly reduces the number of supporting passages while maintaining high Recall.
  % \item We show LLMs are effectively zero-shot NDIs in their ability to generate relevant document identifiers when given only a question.
\end{enumerate}

% The rest of our work is organized as follows. In Section \ref{sec:related_work} we discuss the work most related to ours including traditional approaches and existing work in NDIs.
% In Section \ref{sec:method} we define the problem of retrieval for ODQA and formally introduce our proposed method for retrieval.
% In Section \ref{sec:experiment} we discuss our experimental settings along with the results and analysis of both our method and the existing methods for retrieval and full-stack ODQA.
\section{Related Work}
\label{sec:related_work}
% In this section we discuss the areas of work most related to this one.

\subsection{Traditional Document Retrievers}
\label{subsec:retrievers}

Traditional methods such as TF-IDF and BM25 explore sparse retrieval strategies by matching the overlapping contents between questions and passages \cite{robertson2009probabilistic, chen2017reading,yang2019end}.
DPR \cite{karpukhin2020dense} revolutionized the field by utilizing dense contextualized vectors for passage indexing.
It is first initialized as a pretrained BERT model, then trained discriminatively using pairs of queries and relevant documents, with hard negatives from BM25.
Recent research has improved DPR via better training strategies \cite{xiong2020approximate,qu2021rocketqa,hardqa} and passage re-ranking \cite{mao2021reader,yu2021kg,ju2022grape}.
% Approaches to document retrieval are traditionally broken into two separate categories--dense and sparse.
% Sparse retrievers such as BM25 use sparse query and document embeddings to represent their contents.
% To do this, they use normalized word frequency overlap
However, representations of questions and documents are typically obtained independently in modern dual-encoder dense retrieval models~\citep{karpukhin2020dense,xiong2020approximate}, allowing for only shallow interactions between them~\citep{khattab2021relevance}. 

\begin{figure*}[t]
    \centering
    \includegraphics[width=0.99\textwidth]{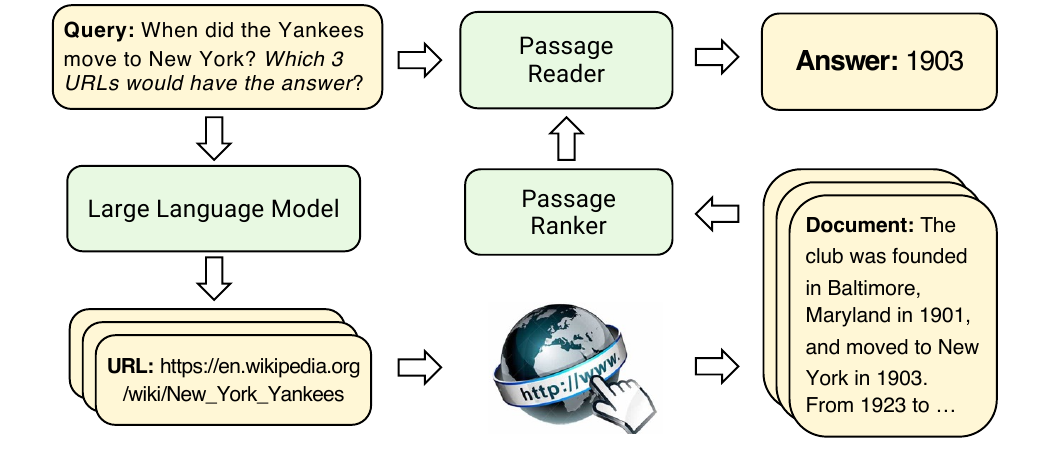}
    \caption{The overall pipeline of our proposed \textsc{LLM-URL}. Given a question, \model first generates a set of URLs which are extracted from the generated text. The URLs are retrieved from the Internet then broken into passages which are ranked and filtered such that only the most relevant are kept. Finally, these passages are given as input to a reader model along with the original question to generate a final answer.}
    \label{fig:diagram}
\end{figure*}

\subsection{Autoregressive Search Engines}
\label{subsec:ndis}

% Scaling up language models in terms of data, compute, and parameters has shown to improve performance on a wide range of downstream tasks \cite{vaswani2017attention, brown2020language}. 
Recent works have investigated the use of auto-regressive language models to generate identifier strings for documents as an intermediate target for retrieval~\cite{yu2022survey}, such as Wikipedia page titles~\citep{de2020autoregressive}, root-to-leaf paths in a hierarchical cluster tree~\citep{tay2022transformer}, or distinctive n-grams that can be mapped to full passages~\citep{bevilacqua2022autoregressive}. 
Since the series of work was carried out almost simultaneously by different research groups, they are often referred to multiple different names in the literature, such as autoregressive search engine, differential search index (DSI), and neural document indexers (NDI). 
% Using Transformers to map directly from question to document identifiers is shown to be an effective alternative to traditional retrieval approaches \cite{tay2022transformer, bevilacqua2022autoregressive, wang2022a}.
Compared to traditional dense document retrievers, these methods leverage a generation model to produce the document indexes.
By forcing the generation model to explain every token in the question and document using cross-attention, the generation abilities of the model significantly improve.
% As the cross-attention step in a generation model is highly expressive, requiring the model to explain every token in the question and document, resulting in a better estimation. 
Our work is closely related to these works, showing experimentally that properly prompting pre-trained large language models can achieve better performance than traditional dense retrieval models~\cite{ouyang2022training,yu2022generate} .

\section{Proposed Method}
\label{sec:method}

In this section we describe a new method, which we refer to as \textsc{LLM-URL}, that employs a large language model (LLM) to perform effective and efficient web document retrieval for knowledge-intensive NLP tasks such as open-domain question answering (ODQA).
% Reviewer 2 didn't like this sentence:
% It is named as \model because it uses the LLM to (autoregressively) generate URLs to find the documents.

ODQA is a two step process consisting of a \textit{retriever} and a \textit{reader}.
Given a question $q$, the goal of the \textit{retriever} is to find the top-$n$ passages $\mathcal{P}_n$ relevant to answering $q$.
Given $q$ and the top-$n$ relevant passages $\mathcal{P}_n$, the goal of the \textit{reader} is to use internal knowledge along with $\mathcal{P}_n$ to generate a correct answer $a$ to question $q$.
The passage retriever plays an essential role in this process.
When $\mathcal{P}_n$ contains more passages that have the correct answer, the reader has a higher chance of finding it.
Instead of \emph{heavily training a dedicated retriever}, our \model solves the problem in a different way as shown in Figure~\ref{fig:diagram}.

Given a question $q$, our \model should find a set of relevant passages to $\mathcal{P}_n$ and give it to the reader.
First, it prompts a LLM (e.g., GPT-3) to directly generate $m$ URLs for $q$. By default, it uses ``Which $m$ Wikipedia URLs would have the answer?'' as the instruction which is appended to each input question as the prompt.
We also append the beginning of the Wikipedia URL (\url{https://en.wikipedia.org/wiki}) to the end of the prompt to encourage the generation of URLs and restrict generation to the Wikipedia article URL format.
As LLM has the ability of in-context learning, we take this advantage to enable the few-shot setting in the prompt. The prompt described above also includes a series of in-context demonstrations.
Each demonstration contains a question sampled from the training set following the prompt described above.
At the end of each demonstration, $m$ URLs which point to gold-labeled documents are listed.
In the zero-shot setting, the original prompt is used without any demonstrations.
In the few-shot setting, the original prompt appended to a series of $d$ demonstrations ($d$=10 in this work).

Given the prompt, the LLM returns a generated sequence of tokens.
Ideally these tokens would construct a sequence of $m$ separated URLs.
In practice, the generated sequence often has extra information such as a proposed answer that is unreliable and needs to be filtered.
We use a regular expression to extract all URLs from the sequence and discard all extra information.
This also filters out many URLs that are improperly formatted.
After extraction, GET requests are made using the extracted URLs and the contents of each retrieval is used to create a set of fetched documents $\mathcal{D}_f$.
Often, $|\mathcal{D}_f| < m$ because some of the generated URLs do not follow a correct format or do not point to real web pages on the Internet.

\begin{table*}
\centering
\resizebox{0.85\textwidth}{!}{
\setlength{\tabcolsep}{1.5mm}{
\begin{tabular}{l|ccccccc}
\toprule
\multirow{2}{*}{Method} & \multicolumn{3}{c}{Document Recall@1} & \multicolumn{3}{c}{Document Recall@10} \\
& WebQ & NQ & TriviaQA & WebQ & NQ & TriviaQA \\ 
% & parameters & open test & open test & wiki split & open test \\  
\midrule
% \multicolumn{6}{l}{\textit{*with retrieved supporting documents from Wikipedia or Google search}} \\
% DPR  & 69.5 & 66.2 & 66.1 & 86.0 & 83.2 & 82.5 \\
Contriever~\cite{izacard2021contriever} & 63.8 & 53.2 & 60.6 & 63.8 & 80.8 & 82.5 \\
BM25~\cite{robertson2009probabilistic} & 49.5 & 47.2 & 63.0 & 81.5 & 76.8 & 82.3 \\
Google API & 61.1 & 55.5 & 51.4 & - & - & - \\
% \midrule
% DSI$^{\text{*}}$ & - & - & - & - & - & - \\
% SEAL$^{\text{*}}$ & - & - & - & - & - & - \\
\midrule
% \multicolumn{6}{l}{\textit{*with retriever, BUT NOT trained on these datasets}} \\
% BM25 + InstructGPT & 20.5 & 53.3 & 16.0 & 78.7 & 65.2 &  \ \ \ \underline{15.7} & 13.7  \\
% Contriever + InstructGPT & 19.1 & 52.4 & 16.8 & 80.4 & \textbf{66.6} &  \ \ \ 15.5 & \underline{14.0} \\
% Google + InstructGPT & \underline{27.8} & \underline{58.7} & \underline{19.9} & \textbf{82.9} & \underline{66.0} &  \ \ \ 14.8 & 13.2 \\
% \midrule
\model (Zero-Shot) & 76.8 & 61.7 & 71.3 & 87.7 & 83.2 & 85.5 \\
\model (Few-Shot) & \textbf{79.7} & \textbf{62.6} & \textbf{73.5} & \textbf{89.9} & \textbf{83.9} & \textbf{86.8} \\
% \model SS-ICL (Ours) & 78.4 & 61.1 & 71.5 & 86.8 & 77.9 & 83.2 \\
\bottomrule
\end{tabular}}}
\caption{Document retrieval as measured by Recall@k. Google API Recall@10 results are left out due to high cost.}
% of \textcolor{red}{MJ: explain why Google API has no Recall@10 results.}}
\label{tab:doc-few-shot}
\end{table*}

\begin{figure}[t]
    \centering
    \includegraphics[width=0.48\textwidth]{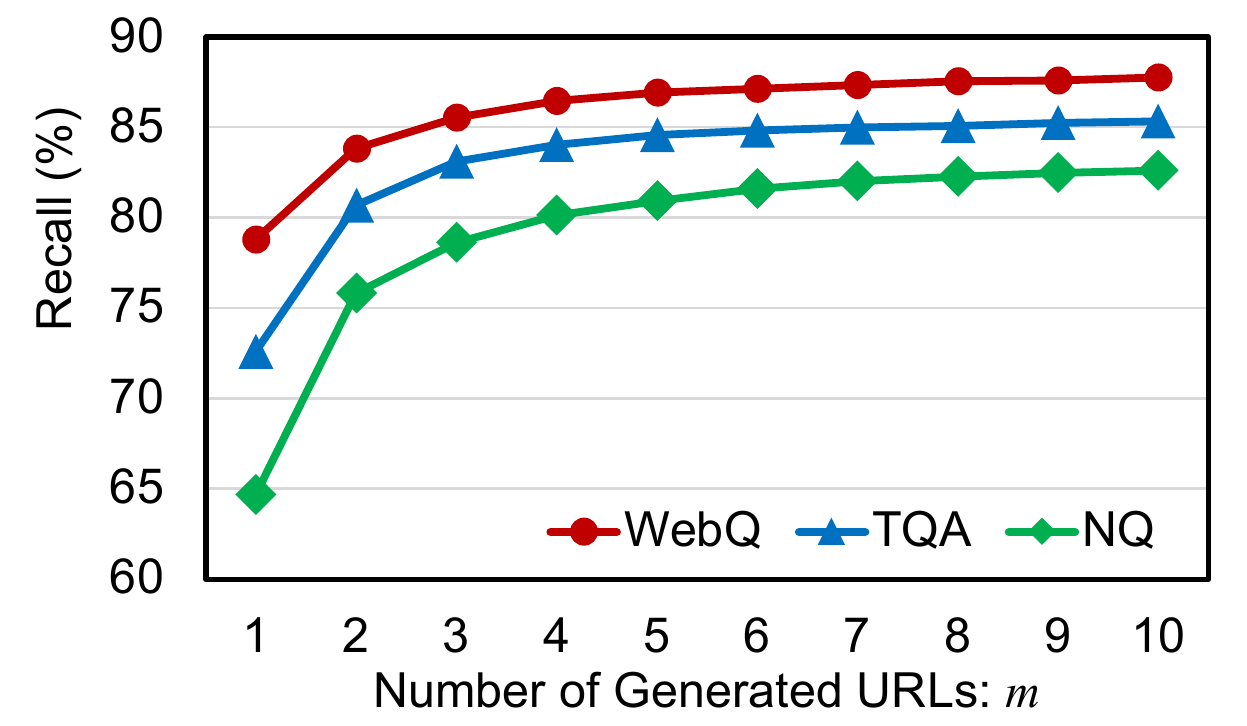}
    \caption{We prompt \model to generate $m$ documents and measure the recall as $m$ increases. Significant recall improvements are seen when $m$ is small but as it increases the marginal benefit decreases.}
    \label{fig:recall_at_k}
\end{figure}

The set of fetched documents $\mathcal{D}_f$ can be passed directly to a reader if $m$ is a small value or the reader being used can handle many large documents.
However, this is usually not the case. Often, $\mathcal{D}_f$ needs to be filtered such that only a small number of the most relevant passages are given to the reader.
To do this, our \model first breaks each document $d \in \mathcal{D}_f$ into a set of small passages.
The passages from each document are collected into a new set, $\mathcal{P}_f$.
A scoring function is used to quantify the relevance of each passage with respect to the question $q$, with high values indicating high relevance with respect to $q$ and low scores indicating low relevance.
A simple scoring function such as BM25 can be used or a more complex one such as DPR \citep{karpukhin2020dense} can.
The passages in $\mathcal{P}_f$ are then sorted from highest to lowest and the top $n$ are kept as $\mathcal{P}_n$.
Finally, $\mathcal{P}_n$ are given to a {reader} along with $q$ to generate an answer.

\paragraph{Advantages of \model:}
Existing autoregressive retrieval methods such as DSI and SEAL use a pre-trained large language model then fine tune it to take questions as input and generate relevant document identifiers as output \cite{tay2022transformer, bevilacqua2022autoregressive}.
Both DSI and SEAL do extensive experiments on a variety of document identifiers which are generated by a heavily trained language model.
Examples of these identifiers include unstructured atomic identifiers, naively structured string identifiers, hierarchical document clustering, and others.
\model instead uses pre-existing document identifiers that exist on the internet: URLs.
Using URLs instead of the aforementioned identifiers has multiple advantages.
URLs often contain words related to the information they link to, allowing for strong association of topics with their URLs.
For example, the title of each Wikipedia page is used in its URL, allowing the LLM is able to directly generate the URL by leveraging semantic information from the question.
To validate the importance of URLs themselves, we also experiment with prompting the LLM to generate Wikipedia titles instead of URLs and find Recall@1 significantly reduces compared to prompting for URL generation.
We believe this is because the URL format itself helps prompt the model for specific information in a specific format.
Further, the use of URLs allows us to simply obtain the evidence document via a HTTP request without any need of training a model or building an index to find the mapping between identifiers and documents.

\section{Experiments}
\label{sec:experiment}

In this section, we present and discuss results from our experiments to ``directly'' demonstrate that our \model is a strong retriever and ``indirectly'' show that it achieves competitive performance on the ODQA task against state-of-the-art solutions.

\paragraph{Large Language Model:}
Following Figure \ref{fig:url-reconstructions}, the large language model we use to generate URLs for our experiments is GPT-3 \textit{text-davinci-003} with greedy decoding and a temperature of 0.
A variety of different prompts are tested for generating URLs, but little difference in performance is observed, so we simply use the best performing prompt which is discussed in Section \ref{sec:method}.

\paragraph{Datasets:} We use three ODQA datasets including Web Questions, Natural Questions, and Trivia QA. We use them to perform evaluation on both the task of document or passage retrieval and ODQA itself.

\subsection{Retrieval}

We expect retrievers to find the most relevant documents and/or passages. We conduct experiments on both document retrieval and passage retrieval.

\begin{table*}
\centering
\scale[0.92]{\begin{tabular}{l|cccccccccc}
\toprule
\multirow{2}{*}{Method} & \multicolumn{3}{c}{Passage Recall@1} & \multicolumn{3}{c}{Passage Recall@10} & \multicolumn{3}{c}{Passage Recall@100} \\
& WebQ & NQ & TriviaQA & WebQ & NQ & TriviaQA & WebQ & NQ & TriviaQA \\ 
% & parameters & open test & open test & wiki split & open test \\  
\midrule
% \multicolumn{6}{l}{\textit{*with retrieved supporting documents from Wikipedia or Google search}} \\
% DPR & 45.4 & 44.6 & 53.2 & 70.5 & 74.5 & 75.3 & 83.1 & 86.5 & 84.8 \\
Contriever & 18.2 & 18.8 & 34.0 & 55.7 & 54.8 & 67.9 & 79.8 & \textbf{79.6} & 83.3 \\
BM25 & 19.1 & 22.8 & 46.2 & 51.8 & 55.6 & 71.7 & 76.6 & \textbf{79.6} & 83.9\\
% Google & - & - & - & - & - & - & - & - & -\\
\midrule
% \multicolumn{6}{l}{\textit{*with retriever, BUT NOT trained on these datasets}} \\
% BM25 + InstructGPT & 20.5 & 53.3 & 16.0 & 78.7 & 65.2 &  \ \ \ \underline{15.7} & 13.7  \\
% Contriever + InstructGPT & 19.1 & 52.4 & 16.8 & 80.4 & \textbf{66.6} &  \ \ \ 15.5 & \underline{14.0} \\
% Google + InstructGPT & \underline{27.8} & \underline{58.7} & \underline{19.9} & \textbf{82.9} & \underline{66.0} &  \ \ \ 14.8 & 13.2 \\
% \midrule
\model (Zero-Shot) & 22.2 & 24.0 & 46.7 & 63.1 & 60.6 & 76.6 & 83.8 & 78.3 & 83.6 \\
\model (Few-Shot) & \textbf{22.3} & \textbf{25.5} & \textbf{49.1} & \textbf{64.8} & \textbf{60.8} & \textbf{77.8} & \textbf{85.9} & 79.0 & \textbf{84.8} \\
\bottomrule
\end{tabular}}
\vspace{-0.1in}
\caption{Passage retrieval as measured by Recall@1, Recall@10 and Recall@100. Here \model is equipped with BM25 to perform the ranking task.}
\label{tab:passage-few-shot}
\vspace{-0.05in}
\end{table*}

\paragraph{Evaluation metrics.} Recall@$k$ ($k$=1, 10, 100) is calculated by measuring the percentage of documents or passages in the top-$k$ which contain one of the gold labeled answers while exact match is calculated by the percentage of predicted answers which match one of the gold labeled answers.
While \model is not constrained by which URLs can be generated for document retrieval, we restrict all generations to Wikipedia URLs only for fair comparison, as discussed in Section \ref{sec:method}
All baseline models also use Wikipedia for retrieval, with some fetching documents in real time and others fetching from an offline corpus.

\subsubsection{Document Retrieval}
\label{subsubsec:document-retrieval}

\paragraph{Baselines:}
Contriever~\cite{izacard2021contriever} and
BM25~\cite{robertson2009probabilistic} are usually used for passage retrieval.
Contriever is a dual encoder which uses a dot product between dense representations of a question and passage to calculate relevance.
BM25 is a sparse retriever which uses the overlapping contents between question and passage to calculate relevance.
Because we use the same passage size to chunk Wikipedia documents, we were able to map their retrieved passages back to the original documents.
We use Google API~\cite{brin1998anatomy} restricted to Wikipedia as a third baseline to retrieve relevant documents given a question.

Existing works such as DSI and SEAL have investigated the use of autoregressive language models to generate identifier strings for documents as an intermediate target for retrieval.
DSI is a Transformer which has been trained to map directly from question to document identifiers by memorizing the contents of the entire corpus \citep{tay2022transformer}.
SEAL is a variant of DSI which uses ngrams as document ids to improve retrieval performance \citep{bevilacqua2022autoregressive}.
Neither DSI nor SEAL report retrieval results on full documents and do not have publicly available implementations, so they are left out and discussed in Table~\ref{tab:passage-ndi} and Section~\ref{subsubsec:passage-retrieval} on passage retrieval.

Unlike the baselines, our \model employs an LLM.
It has two settings: zero-shot and few-shot.
In the zero-shot setting, no in-context demonstrations are given whereas in the few-shot setting a few demonstrations are appended to the prompt.

\begin{figure}[t]
    \centering
    \includegraphics[width=0.48\textwidth]{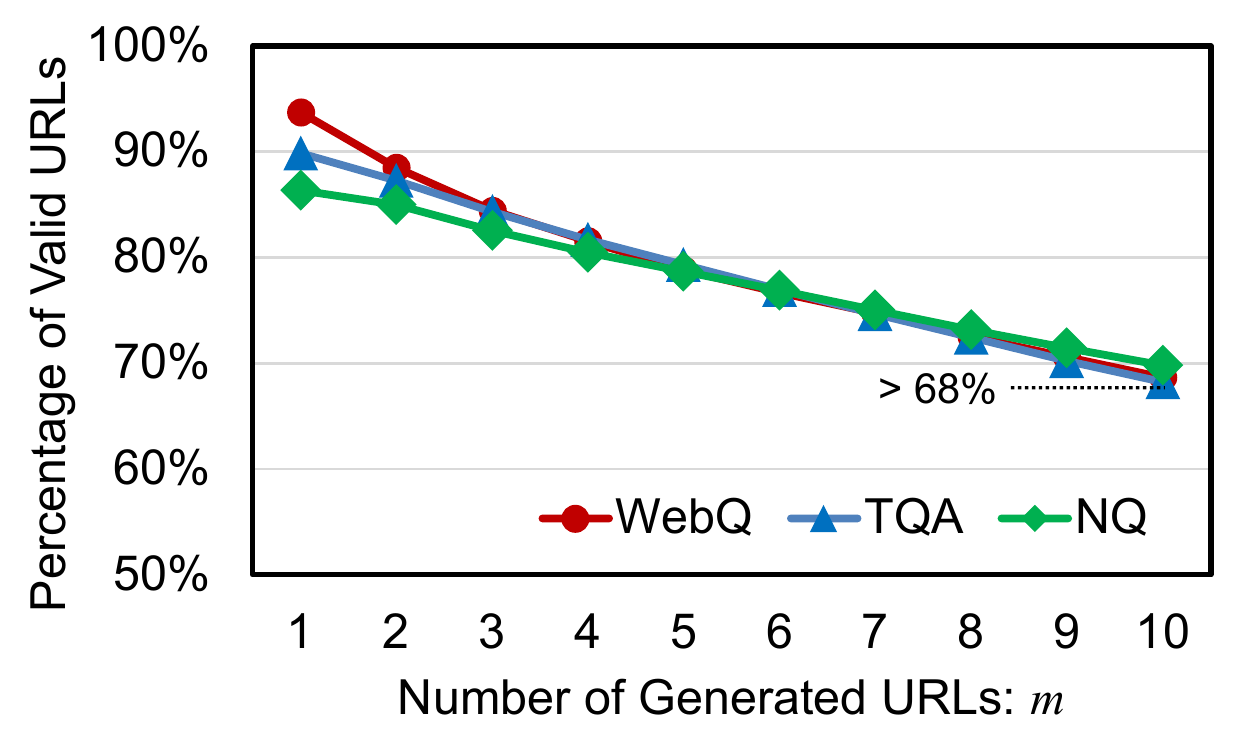}
    \vspace{-0.3in}
    \caption{The percentage of valid URLs generated from \model as the total number of generated URLs $m$ increases from 1 to 10. As $m$ increases, invalid URL generations become more frequent.}
    \label{fig:valid-generations}
    \vspace{-0.1in}
\end{figure}

% In this section, we present and discuss results from our experiments on three different ODQA datasets including Web Questions, Natural Questions, and Trivia QA.
% Our experiments can largely be broken into two categories: retrieval and downstream question answering.
% The evaluation metrics we use for retrieval and downstream question answering are recall@$k$ and exact match (EM) respectively.

\paragraph{Results:} The results of our document retrieval experiments are shown in Table~\ref{tab:doc-few-shot}.
In this setting Recall@$k$ is calculated directly after the documents are retrieved with no intermediary steps.
\model significantly outperforms baseline methods on all datasets for both Recall@1 and Recall@10.
Specifically, zero-shot \model improves document Recall@1 relatively by 20.4\%, 11.2\%, and 13.2\% over the strongest baseline on WebQ, NQ, and TriviaQA, respectively. Few-shot \model further expands the improvement to 24.9\%, 12.8\%, and 16.7\%, respectively. URLs can be extracted from the large-scale parameters of LLMs, and these URLs can lead to more accurate documents than what existing methods can retrieve.
Both the LLM parameters and in-context demonstrations are significantly useful in document retrieval.

Figure~\ref{fig:recall_at_k} shows Recall scores converge when the number of generated URLs $m$ increases. Due to the diminishing returns from increasing $m$, our experiments do not explore values of $m$ greater than 10.

\begin{table}
\centering
\setlength{\tabcolsep}{1mm}{
\resizebox{0.47\textwidth}{!}{
\begin{tabular}{l|ccc}
\toprule
\multirow{1}{*}{Method} & \multicolumn{1}{c}{Recall@1} & \multicolumn{1}{c}{Recall@10} \\
% & NQ & NQ\\
% & parameters & open test & open test & wiki split & open test \\  
\midrule
DSI$^{\text{1}}$ & 25.1 &  56.6 \\
SEAL$^{\text{1}}$ & \textbf{26.3} & \textbf{74.5} \\
% \multicolumn{2}{l}{$^{\text{2}}$\textit{not zero-shot}} \\
\midrule
% \multicolumn{6}{l}{\textit{*with retriever, BUT NOT trained on these datasets}} \\
% BM25 + InstructGPT & 20.5 & 53.3 & 16.0 & 78.7 & 65.2 &  \ \ \ \underline{15.7} & 13.7  \\
% Contriever + InstructGPT & 19.1 & 52.4 & 16.8 & 80.4 & \textbf{66.6} &  \ \ \ 15.5 & \underline{14.0} \\
% Google + InstructGPT & \underline{27.8} & \underline{58.7} & \underline{19.9} & \textbf{82.9} & \underline{66.0} &  \ \ \ 14.8 & 13.2 \\
% \midrule
\model (Zero-Shot) & 24.0 & 60.6  \\
\model (Few-Shot) & 25.5 & 60.8 \\
\bottomrule
% \multicolumn{2}{l}
\multicolumn{2}{l}{$^{\text{1}}$\textit{explicitly trained for retrieval on NQ}} \\
\end{tabular}}}
\caption{
Passage retrieval as measured by Recall@1 and Recall@10. \model is equipped with BM25 for passage ranking.
Other datasets are left out due to not being reported in either paper and no public implementations.}
\label{tab:passage-ndi}
\vspace{-0.1in}
\end{table}

\paragraph{Are the generated URLs valid?} It is worth noting that the generated URLs are not always valid.
Some generated URLs do not have valid URL syntax and some point to Wikipedia pages that do not exist.
Rarely, URLs will be generated for domains aside from Wikipedia.
For fair comparison, all of these faulty URLs are discarded and only documents coming from valid Wikipedia articles are kept.

Further analysis is done to measure the ratio of valid Wikipedia URLs while the total number of generated URLs $m$ increases from 1 to 10, shown in Figure~\ref{fig:valid-generations}.
The number of valid URL generations remains surprisingly high (i.e., higher than 68\%) as $m$ increases from 1 to 10. However, the rate of valid generations appears to fall off as $m$ increases, indicating there are diminishing returns from each marginal increase of $m$.
% The ratio decreases, because the quality becomes less reliable, but remains higher than 68\% consistently on the three datasets.

% Fair analysis between \model and existing baselines remains difficult due to the unique few-context setting, which existing baselines do not have.
% To remedy this we focus our experiments and analysis on the few-shot setting while supplementing all results with the zero-shot setting where appropriate for fair comparison.

% While \model by itself can only fetch full documents, few downstream readers can take full documents as input due to their long length.
% However, if combined with the passage ranking mechanism described in Section \ref{sec:method} it effectively becomes a passage retriever.
% Therefore our retrieval experiments can be largely split into two categories: document retrieval and passage retrieval.

% In both settings, URLs are extracted directly from the LLM generations and retrieved from the Wikipedia API via GET requests.
% Often, predicted answers to the original question are generated by the LLM but these and all other generation artifacts which do not contain URLs are ignored.
% These diminishing returns can be further seen in Figure \ref{fig:recall_at_k} which shows the recall as $k$ increases instead of the number of valid URLs generated.
% It is worth noting as well that increasing $m$ significantly increases the computation required for retrieval.

\subsubsection{Passage Retrieval}
\label{subsubsec:passage-retrieval}

\paragraph{Baselines:} Four methods, including Contriever, BM25, DSI \citep{tay2022transformer}, and SEAL \citep{bevilacqua2022autoregressive}, were introduced in Section~\ref{subsubsec:document-retrieval}. Google API was used for document retrieval and not applied to passages.

\paragraph{Results:} The results of our passage retrieval experiments are shown in Table \ref{tab:passage-few-shot}.
In this setting Recall@$k$ is calculated on the top-$k$ passages ranked by the ranker instead of just on the raw documents shown in Table \ref{tab:doc-few-shot}.
\model performs slightly better than baseline methods for Recall@1 and Recall@10 and as well as baseline methods for Recall@100.
In the zero-shot setting, \model improves relative Recall@1 by 16.2\%, 5.3\%, and 1.1\% with respect to the strongest baseline on WebQ, NQ, and TriviaA respectively. The few-shot setting of \model expands the improvement to 16.8\%, 11.8\%, and 6.3\%, respectively. For Recall@10, similar improvements can be seen.

\begin{table}
\centering
\resizebox{0.47\textwidth}{!}{
\setlength{\tabcolsep}{0.6mm}{
\begin{tabular}{l|ccc}
\toprule
\multirow{2}{*}{Method} & \multicolumn{3}{c}{Zero-Shot QA EM} \\
& WebQ & NQ & TriviaQA \\ 
% & parameters & open test & open test & wiki split & open test \\  
\midrule
% \multicolumn{6}{l}{\textit{*with retrieved supporting documents from Wikipedia or Google search}} \\
% DPR + InstructGPT & 20.1 & 30.0 & 55.3 \\
Contriever + InstructGPT & 16.8 & 19.1 & 52.4 \\
BM25 + InstructGPT & 16.0 & 20.5 & 53.3 \\
Google + InstructGPT & 19.9 & 27.8 & 58.7 \\
GenRead (InstructGPT) & 24.8 & 28.2 & 59.3 \\
\midrule
DSI$^{\text{1}}$ + FiD & - & ~ 31.4$^{\text{2}}$ & - \\
SEAL$^{\text{1}}$ + FiD & - & \textbf{43.6} & 41.8 \\
\midrule
InstructGPT (no docs.) & 18.6 & 20.9 & 52.6 \\
\model (Zero-Shot) & 28.1 & 26.4 & 60.1 \\
\model (Few-Shot) & \textbf{29.0} & 27.3 & \textbf{60.7} \\
% \model + FiD  & 33.9 & 29.8 & 56.9 \\
\bottomrule
\multicolumn{4}{l}{$^{\text{1}}$\textit{explicitly trained for retrieval on NQ}} \\
\multicolumn{4}{l}{$^{\text{2}}$\textit{result from \citet{bevilacqua2022autoregressive}}} \\
\end{tabular}}}
\caption{Zero-shot open-domain QA performance as measured by exact match (EM). All \model models use InstructGPT as the reader unless otherwise stated.}
\label{tab:downstream-qa}
\vspace{-0.1in}
\end{table}

% For Recall@10, similar improvements can be seen.
% In the zero-shot setting, \model improves the relative Recall@10 by 13.3\%, 9\%, and 8.5\% with respect to the strongest baseline on WebQ, NQ, and TriviaQA respectively. Again, the few-shot setting of \model expands the improvement to 16.3\%, 9.4\%, and 9\%, respectively.

For Recall@100, performance is better relative to baseline models for all datasets except NQ.
In the zero-shot setting, \model improves the relative Recall@100 by 5.0\% for WebQ and performs slightly worse than the best baseline method on NQ and TriviaQA by 1.7\% and 0.4\% respectively. The few-shot setting of \model for Recall@100 shows a slight improvement on WebQ and TriviaQA,
% by 7.6\% and 1\% respectively,
but performs slightly worse than the strongest baseline on NQ.
% by 0.8\%.

Despite being limited to only the passages from 10 documents, \model performs better than baseline methods for smaller $k$ and performs as well as baseline methods for higher values of $k$.
% Our method performs slightly better than Contriever and BM25 for low values of $k$, but outperforms all baseline methods for larger values of $k$. \textcolor{red}{MJ: [is this sentence true or false?] \nz{this was true before we had results for Recall@100 this morning...}}

% The results of our zero-shot passage retrieval experiments are shown in Table \ref{tab:passage-few-shot}.
% In this setting, the full documents retrieved from the generated URLs are broken into small passages which are then searched using a ranker such as BM25.
% For more details on this setting, see Section \ref{sec:method}.
% Recall@$k$ is then calculated on the top-$k$ passages ranked by the ranker.
% Our method performs slightly better than Contriever and BM25 for low values of $k$, but outperforms all baseline methods for larger values of $k$.

% \subsubsection{NDI Passage Retrieval}
% \label{subsubsec:ndi-passage-retrieval}

% \textcolor{red}{MJ: [introduce Table 3]}
The comparison between \model and existing document identifier-based methods such as DSI and SEAL are shown in Table \ref{tab:passage-ndi}.
% Performance calculations remain the same in this setting, with Recall@$k$ being calculated on the top-$k$ passages ranked by the ranker.
For Recall@1, zero-shot \model performs slightly worse than the best baseline by 8.8\%.
This performance gap is slightly smaller in the few-shot setting with \model performing 3.1\% worse than the best baseline.
For Recall@10, zero-shot \model performs worse than the best baseline by 18.7\%.
Few-shot \model performs only slightly better than the zero-shot setting, performing worse than the best baseline by 18.4\%.

% Despite never being explicitly trained for retrieval on NQ, \model performs only 

% Notably, \model achieves roughly the same performance as DSI and SEAL despite never being trained explicitly for retrieval.
% Neither DSI nor SEAL report retrieval results for Recall@100 or for WebQ or TriviaQA, and no publicly available implementations are provided, so these results are left out.

% \subsection{Few-Shot Setting}
% Here we discuss the performance of \model in the few-shot setting using in-context learning.
% We add demonstrative exemplars in the prompt to the LLM. To be specific, the LLM is prompted with a series of questions (exemplars) with each question followed by 10 Wikipedia URLs that likely contain the answer.
% The LLM is then prompted with a new question from the dataset and asked to generate $m$ URLs that would have the answer.
% The rest of the retrieval pipeline remains the same.
% The few-shot setting improves performance of standard \model. This shows that adding more demonstrative exemplars helps improve the retrieval accuracy of \model. However, the margin between zero-shot setting and few-shot setting is not huge, indicating that the LLM already has the ability to generate relevant URLs itself. Besides, we empirically find that the few-shot setting leads to fewer non-existent URLs than zero-shot setting, which should contribute to the gold URLs presented in the exemplars.

\subsection{Open-Domain Question Answering}

\paragraph{Evaluation metric:} We use exact match (EM), which is short for \textit{exact string match with the correct answer}, because the goal of ODQA is to find an exact answer to any question using Wikipedia articles.

\paragraph{Results:}
Here we discuss the downstream QA performance of \textsc{LLM-URL}.
In this setting, an answer only has an exact match if the normalized generated text is within the list of acceptable answers to a question.
When combined with InstructGPT as a reader, \model performs significantly better on WebQ and slightly better on TriviaQA when compared with the best performing baseline methods.
On NQ, \textsc{LLM-URL}+InstructGPT performs worse than baseline NDIs and only slightly worse than the best remaining baseline.
In the zero-shot setting, \textsc{LLM-URL}+InstructGPT improves upon the best baseline method by 13.3\% and 1.3\% on WebQ and TriviaQA respectively.
\model+InstructGPT performs worse than the best baseline method by 39.5\% on NQ.
In the few-shot setting, \textsc{LLM-URL}+InstructGPT performs better than the best baseline method by 16.9\% and 2.3\% on WebQ and TriviaQA respectively.
\textsc{LLM-URL}+InstructGPT performs worse than the best baseline method by 37.4\% on NQ.

Despite not being explicitly trained for retrieval, \textsc{LLM-URL}+InstructGPT performs significantly better than baseline methods for WebQ, achieves on-par performance with existing methods for TriviaQA, and performs slightly worse than existing methods for NQ.

Our results indicate \model could be a promising solution to retrieval for a wide range of knowledge intensive tasks with little to no training data required.
% When combined with a generator (e.g., Transformer decoder), it can be used to improve factual summarization, question generation, or scientific text generation among others.

\subsection{Discussions}

\subsubsection{Time Sensitive Queries}

There are a number of additional qualitative benefits that \model has over existing methods.
One large advantage of \model is that the documents are retrieved in real time from the source.
So long as the source stays up to date without the URL itself changing, our proposed method is capable of answering time sensitive queries without any extra modifications.

In contrast, existing dual encoder approaches such as Contriever require a document to be re-encoded each time it changes.
Existing methods such as SEAL \citep{bevilacqua2022autoregressive} and DSI are also tricky to keep up to date for time sensitive queries as the LLM would have to be retrained to learn the new content of the updated documents.

\subsubsection{Frequent Entities Analysis}
\begin{figure}[t]
    \centering
    \includegraphics[width=0.48\textwidth]{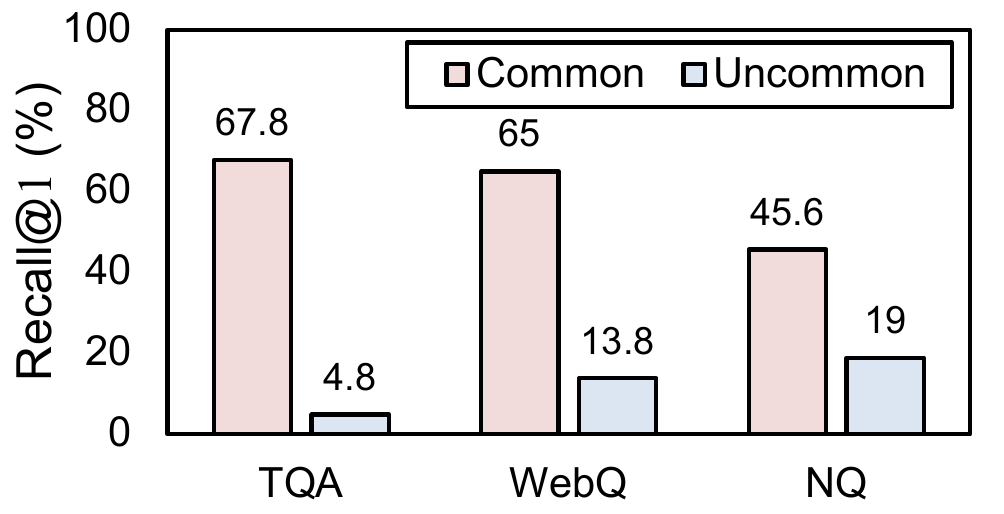}
    \vspace{-0.3in}
    \caption{Common vs Uncommon entity recall@1. Common is defined as a question containing entities in the top-1 million most common entities on Wikipedia. \model performs much better when retrieving information about common entities.}
    \label{fig:freq-analysis}
    \vspace{-0.1in}
\end{figure}

\begin{table*}[ht]
\centering
\setlength{\tabcolsep}{1.5mm}{
\resizebox{\textwidth}{!}{
\begin{tabular}{l|cc||l|c||l|c}
\toprule
\textbf{\model} & Exists & Answer & \textbf{Contriever} & Answer & \textbf{BM25} & Answer\\
\midrule

\url{wiki/Jellyfish} & \cmark & \cmark & Smack (ship) & \xmark & Collective noun & \xmark \\
\url{wiki/Collective_noun} & \cmark & \xmark & Collective noun & \xmark & Determiner & \xmark \\
\url{wiki/Smack_(group)} & \xmark & \xmark & Cetacean intelligence & \xmark & Glass sea creatures & \xmark \\
\url{wiki/Cnidaria} & \cmark & \cmark & Well smack & \xmark & Minotaur & \xmark \\
\url{wiki/Medusozoa} & \cmark & \cmark & Plankton & \xmark & Mass noun & \xmark \\
\url{wiki/Scyphozoa} & \cmark & \cmark & Sperm whale & \xmark & Well smack & \xmark \\
\url{wiki/Cubozoa} & \cmark & \cmark & Loaded question & \xmark & Nomenclature & \xmark \\
\url{wiki/Hydrozoa} & \cmark & \cmark & Jabberwocky & \xmark & Archomental & \xmark \\
\url{wiki/Staurozoa} & \cmark & \cmark & Merrow & \xmark & Honey Smacks & \xmark \\
\url{wiki/Rhizostomeae} & \cmark & \cmark & Loaded question & \xmark & Well smack & \xmark \\

\bottomrule
\end{tabular}}}
\caption{Case study of retrieved documents from the question ``\textit{A 'smack' is a collective noun for a group of which sea creatures?}''. ``Exists'' means whether the URL points to a valid Wiki page. ``Answer'' means whether the document contains the answer. We omit the prefix of generated URLs for brevity (\url{https://en.wikipedia.org/}). For BM25 and Contriever, we show the document titles of the top-10 retrieved passages, respectively. The correct answer is ``jellyfish.''}
\label{tab:case-study}
\end{table*}

Following \cite{mallen2022when}, we analyze the retrieval performance of \model when the gold-labeled answer entity is common versus when it is not.
For each question-answer pair in a given dataset we check to see if the labeled entity exists within the top one-million common entities from Wikipedia.
Using this, we split our dataset into two distinct subsets: question-answer pairs that contain a common entity and those that do not.
In measuring the performance of our model on these two distinct sets across Web Questions, Natural Questions, and TriviaQA, we find \model performs significantly better on common entity question-answer pairs.
The results of our analysis are shown in Figure~\ref{fig:freq-analysis}.
Across all three datasets, the recall of common-entity question-answer pairs is many times greater than the recall from the rest of the dataset.

Previous work has shown LLMs in the closed-book setting, where the model must rely solely on the information contained within its weights, perform much better on common-entities versus uncommon ones \citep{mallen2022when}.
Our results show this problem extends beyond the closed-book setting and also applies to retrieval when using \textsc{LLM-URL}.
This also could explain the high word count from documents we found when evaluating \textsc{LLM-URL}.
The average Wikipedia article is 644 words, but the average word count from Wikipedia documents retrieved via \model was 10k.
We believe this discrepancy is caused by common entities having much more detail in their Wikipedia articles and in turn having much higher word count.

% \subsubsection{Prompts Analysis}
% \label{subsubsec:prompts}
% We experiment with a wide variety of prompts for evaluation of \model for both the few-shot and zero-shot setting.
% The simplest of our prompts is ``{{question}} Which Wikipedia URL has the answer?''
% For this simple base-case we also try, ``{{question}} Which Wikipedia URL would have the answer?'' and ``{{question}} Here is a Wikipedia article I found that has the answer''.
% Of these, all performed nearly identically, so the first one was used for all experiments.

% However, these prompts only generate a single URL.
% To prompt the model to generate multiple URLs, we simply prompt with, ``\{\{question\}\} Which \{\{$k$\}\} Wikipedia URLs would have the answer?''
% In this case, the LLM often generates an ordered list of 10 URLs.

% For the few-shot setting, we simply repeat the prompt above with questions from a training set and URLs which contain the gold-labeled answer.
% After the in-context demonstrations are given, the prompt is given again with the intended question and a URL is not supplied.

% Here is a simple example prompt for retrieving a document from two in-context demonstrations:
% \nz{Is there a good way I can make this a \fbox{text block}?}
% "What do Jamaican people speak? Which Wikipedia URL has the answer?

% \url{https://en.wikipedia.org/wiki/Jamaican_Creole}

% What did James K Polk do before he was president? Which Wikipedia URL has the answer?

% \url{https://en.wikipedia.org/wiki/James_K._Polk}

% \{\{question\}\} Which Wikipedia URL has the answer?"

\subsubsection{Case Study}
\label{subsubsec:casestudy}

In Table~\ref{tab:case-study}, we show a case study comparing \model with two baseline retrieval methods, BM25 and Contriever, on the question ``A `smack' is a collective noun for a group of which sea creatures?'' which is in the TriviaQA test set.
The gold-labeled answer to this question is ``jellyfish''.

In the closed-book setting, InstructGPT mistakenly predicts ``dolphins'' as the answer.
When using Contriever to retrieve 10 passages from Wikipedia given the query, none of the passages contains the gold answer.
For instance, Contriever retrieves passages about ``smack'', a kind of fishing vessel, along with other passages about sperm whales, plankton, and other unrelated topics.
Similar results are found while using BM25 as the retriever.

In contrast, \model performs much better in this scenario, retrieving 7 documents which contain the answer. The top retrieved document is exactly about the gold answer ``jellyfish''. The fourth to the tenth documents all talk about different types of jellyfish. After being chunked into passages then sorted by the ranker, the top 10 passages are concatenated.
Among them, it contains ``A group of jellyfish is called a smack,'' which contains the answer to the question and comes directly from the first retrieved document, titled ``jellyfish.'' When InstructGPT is then prompted with these 10 passages along with the question, the gold answer ``jellyfish'' is correctly generated.

This case study highlights multiple advantages of \model.
First, \model finds documents related to both the question and the answer.
It directly locates documents that talks about ``jellyfish'' instead while BM25 and Contriever locate documents related to the question only--not the answer.
Second, \model is more precise than BM25 or Contriever.
In this case, 7 out of 10 generated URLs from \model point to a Wikipedia document that contains the answer.
However, both BM25 and Contriever fail to retrieve any documents containing the answer.
Third, the set of documents retrieved by \model are complementary to each other, while in BM25 or contriever, each document in the top-10 is selected independently.
This is because the LLM is able to refer to previous generated URLs before it generates the next one, allowing each newly generated URL to be conditioned on all the previous URLs.
This leads to a more informative evidence context in open-domain question answering.

% Third, \model has built-in conditional retrieval.
% If there are multiple relevant topics for a question, once \model generates a URL for one topic it will next generate a URL for a completely different yet complimentary topic.
% This cycle continues until \model has exhausted all relevant topics.

% Note that after the 3rd URL is generated, the rest of the URLs are different types of jellyfish.

% \input{5-future}
\section{Conclusion and Future Work}
\label{sec:conclusion}

In this paper, we explored whether large language models can generate URLs prompted by human instructions for document retrieval.
Surprisingly, we found that by providing a few (query, URL) pairs as in-context demonstrations, large language models (e.g. GPT-3) generated Web URLs where near 90\% of the corresponding documents contain correct answers to open-domain questions in WebQ.
% In this work we show \model is a simple and effective approach to document retrieval which borrows many of the appealing properties of NDIs while leveraging the scale of LLMs.
% We show LLMs such as GPT-3 have strong capabilities of generating valid URLs for documents that contain the answer to a given question.
Furthermore, by breaking the retrieved documents into passages then ranking them with BM25, we showed a significant number of unnecessary passages could be filtered out while retaining high recall, which outperformed baseline methods by a significant margin.

% \section{Future Work}
% \label{sec:future}

There are numerous exciting directions for future work.
While a number of broad spectrum retrieval benchmarks such as BIER \cite{thakur2021beir} exist, it remains to be seen whether the few-shot demonstrations shown in this work can be further tuned for specific retrieval tasks.
Promptagator \cite{dai2022promptagator} shows significant performance improvements can be achieved by tuning prompts in a similar way.

Further, it remains to be seen whether fine tuning the prompt for each individual question can further improve the retrieval performance.
As with Promptagator, prior work has shown using clustering to select diverse demonstrations for any given question further improves retrieval performance as well as downstream QA performance.
% \cite{yu2022generate}.

\section*{Limitations}
\label{sec:limitations}

Despite the strong performance on the presented datasets, our approach is limited in its ability to update knowledge state and adapt to new domains. 
A major feature of \textit{retrieve-then-read} is the ability to swap in new documents when new information is learned, such as temporally more recent documents, or adding in documents from a new domain to quickly adapt to a new downstream task. 
Our approach relies on a large language model to contain all this knowledge and adding new knowledge would likely require some retraining. 
% Future work will explore how to efficiently incorporate new knowledge into our method.
In addition, large generation models still suffer from hallucination errors, resulting in incorrect predictions.
When tasked with generating 10 URLs, \model may only generate 6 or 7 which link to valid documents.
Finally, our approach involves very large language models, slow web requests, and document processing which may make it cumbersome to use in practice.
\section*{Acknowledgements}
This work was supported by NSF IIS-2119531, IIS-2137396, IIS-2142827, CCF-1901059, and ONR N00014-22-1-2507. Wenhao Yu was partly supported by the Bloomberg Data Science Fellowship.

\nocite{koala}
\nocite{multi-task-survey}

% \clearpage
\balance
\bibliography{bibliography}
\bibliographystyle{style/acl_natbib}

\appendix
% \appendix

% \section{Example Appendix}
% \label{sec:appendix}

% This is an appendix.

\end{document}